\pdfoutput=1
\documentclass{article}
\usepackage{ijcai18}

\usepackage[table]{xcolor}
\usepackage{times}
\usepackage{xcolor}
\usepackage{soul}
\usepackage[utf8]{inputenc}
\usepackage[small]{caption}
\usepackage{graphicx}
\usepackage{booktabs}
\usepackage{multirow}
\usepackage{rotating}

\usepackage{latexsym} 

\title{Analyzing Hypersensitive AI: Instability in Corporate-Scale Machine Learning}

\author{
Michaela Regneri$^\Diamond$,
Malte Hoffmann$^\Diamond$,
Jurij Kost$^\Diamond$,
Niklas Pietsch$^{\bullet}$,
Timo Schulz$^\ast$,
Sabine Stamm$^{\bullet}$
\\ 
$^\Diamond$OTTO, $^\ast$ITGAIN, $^{\bullet}$Otto Group Solution Provider \\
Hamburg, Germany\\
$^\Diamond${\{firstname.lastname\}@otto.de},
$^\ast${timo.schulz@itgain.de},
$^{\bullet}${\{firstname.lastname\}@ottogroup.com}
}

\begin{document}
\definecolor{lgray}{gray}{0.95}
\maketitle

\begin{abstract}
Predictive geometric models deliver excellent results for many Machine Learning use cases. Despite their undoubted performance, neural predictive algorithms can show unexpected degrees of instability and variance, particularly when applied to large datasets. We present an approach to measure changes in geometric models with respect to both output consistency and topological stability. Considering the example of a recommender system using word2vec, we analyze the influence of single data points, approximation methods and parameter settings.
Our findings can help to stabilize models where needed and to detect differences in informational value of data points on a large scale.
\end{abstract}
 
\section{Introduction}\label{introduction}
Nowadays, Artificial Intelligences (AIs) beat humans in recognizing hand-written numbers \cite{handwritings}, playing complex games like Go! \cite{silver2017mastering} and partially even in cancer diagnosis \cite{stanford:skinCancer}. Many algorithms become more performant and more necessary than ever with the steady growth of available data. On the one hand, automated decisions become more accurate with more information to learn from, on the other hand, humans cannot keep pace with the velocity at which new data arises.

With the advance of neural networks and Big Data sources, results and mechanisms of Machine Learning components seem to have become less explainable. 
AI enables us to access useful data in the first place. However, the complexity of AI algorithms increases the gap between users as information consumers and the full background knowledge about the digested input's formation. AI consumers thus cannot always verify reliability and objectivity of automatically compiled content. 

Instability in Machine Learning algorithms often exacerbates the problem of classifying AI output as trustworthy or questionable: The output of a learning algorithm might sometimes alter unexpectedly, even if the environment changes only marginally or not at all.
Such instabilities can often be misinterpreted as unreliability of an algorithm, in particular for systems which require a high degree of guaranteed behaviour and reproducibility (e.g. a diagnostic assistant).

We take a step towards explaining output changes in geometric models in general, including neural models in particular. In an innovative experimental setup for sensitivity analysis, we apply topological and application-oriented metrics to detect differences between two recommender models. The experimental design disentangles two common sources of instability: We separate sensitivity that originates from minimal changes in the input data from variation originating from the algorithm itself, in particular from strategies that keep the approach compuationally tractable for large-scale applications.

Our main contribution is as follows: We present a combination of \emph{scalable measures for sensitivity}, drawn from both, output consistency and topological model change. 
We apply those metrics to analyze word2vec \cite{DBLP:journals/corr/abs-1301-3781}, a widely used example for geometric models. As a result, we classify \emph{factors of instability} originating from the algorithm itself and from properties of data points.

\section{Related Work}\label{related}
While the algorithm in our focus was designed for Natural Language Processing, there are many non-linguistic applications of related vector models. For instance, skip-gram algorithms exist for different types of data, like social media graphs \cite{Perozzi:2014:DOL:2623330.2623732}, ordered products in e-commerce \cite{DBLP:journals/corr/abs-1803-02349}, or click-through data as in our case.

Plenty of previous work is based on word2vec \cite{DBLP:journals/corr/abs-1301-3781},
and its results are analyzed thoroughly, but only few studies explain instabilities and the impact of randomness on the learned vector model. \citeauthor{Wendlandt18Surprising} [\citeyear{Wendlandt18Surprising}] show that even the embeddings of frequent words are unstable. Semantic properties of natural language (in particular, semantic relations) account for some of the sensitivities they find, which are thus hard to generalize to use cases that operate on other data sources rather than text.

\citeauthor{TACL1202} [\citeyear{TACL1202}] show that inherent relations in the input data (like aforementioned semantic relations) can only partially explain the learned vector model. Some of the instability factors they find (like document length) might be use case generic. Other work \cite{DBLP:journals/corr/abs-1804-04212} highlights that instability can be also caused by default hyper-parameters that are not tuned for a specific task.

One evaluation methodology we use resembles the lexical substitution task \cite{McCarthy:2007:STE:1621474.1621483}, which is a popular benchmark for automatically generated semantic similarity measures. Similar embedding methods to the one we employ have also been used for finding lexical substitutions \cite[among others]{DBLP:conf/naacl/MelamudLD15}. In context of recommender systems, the Netflix challenge \cite{Bennett07thenetflix} is a similar setting for predicting user-specific ratings.

\citeauthor{DBLP:journals/corr/abs-1802-04443} [\citeyear{DBLP:journals/corr/abs-1802-04443}] derive a topological metric for measuring data complexity. One part of this metric is the number of the connected components, which we also employ as a metric. However, our goal differs from \citeauthor{DBLP:journals/corr/abs-1802-04443} since they derive the appropriate neural architecture from the input data's topology, whereas we analyze the resulting embeddings.

\section{Data and Algorithm}\label{basics}
In this section, we briefly sketch recommender systems, introduce word2vec and describe our source data.
\subsection{Recommender Systems}
Recommender systems \cite{Resnick:1997:RS:245108.245121} use automated means that help users make decisions with insufficient information. In general, such systems either use collaborative filtering, content-based recommendations or hybrid approaches. Collaborative filtering means that the (implicit or explicit) recommendations of all users are aggregated to find the best recommendations for the current user (such algorithms are mostly employed in web-shops for unknown users). Content-based recommendations refer to products previously bought or viewed by the current user and generate recommendations based on similarities with those products. Such similarities can be visual, textual or computed from other metadata (such as prices or brands).

Like many modern recommenders, the system we consider is a hybrid approach. Given a \emph{seed product} (the product currently viewed by the user), we compute a sorted list of recommendations that are possible \emph{alternatives} for the seed product (\emph{``You might also like...''}). 
Such item-to-item recommendations are widely used in e-commerce. They increase conversion rates by helping customers find a product that meets their current need, and by keeping them engaged on the site.
Technically, the alternatives in our model share similar click embeddings with the seed product, i.e. other customers looked at the alternative in similar user journey contexts in which they visited the current product in focus. For computing such embeddings, we use the word2vec algorithm.

\subsection{Word2vec}
Word2vec \cite{DBLP:journals/corr/abs-1301-3781} is an algorithm that computes word embeddings. Words are represented by vectors, and the computed embeddings represent the words in their predicted contexts. In contrast to count-based co-occurrence models, word2vec and other predictive models use neural networks to compute a vector space, deriving abstract concepts as dimensions for each word from a given corpus. Count-based models define the vector space by directly using context words as dimensions, resulting in many more dimensions that are potentially less effective. Neural models perform superior in numerous tasks \cite{marcobaroni2014predict}.
 
Word2vec applications comprise mostly tasks around Natural Language Processing, e.g. Sentiment Analysis \cite{DBLP:journals/corr/Liu17b}, Information Retrieval \cite{Onal2017NeuralIR} and Named Entity Recognition \cite{Das2017NamedER}, among many others. Neural predictions deliver good quality, which, however, comes at the cost of explainability. The features of word2vec are mostly opaque, and the resulting models change to a surprising extent even with only minor input variations \cite{Wendlandt18Surprising}. The combination of high performance, frequent use,  instability and unpredictability forges high interest in  conditions that influence the algorithm, especially in corporate contexts.

\subsection{Source Data}
Our training data consists of anonymous user interactions with product pages. We define the chronologically sorted series of clicks by one  user with breaks shorter than a certain time a \emph{user session}. While a user session contains events of various types, we only consider clicks on product pages. Adopting linguistic terminology, we call one click in a session a \emph{token}, and one abstract product, that can have multiple click instances, a \emph{(product) type}. We use the term \emph{product} interchangeably with \emph{type}.   The vector model we compute contains one vector per product type.

In our data model, each product belongs to a \emph{product group} (e.g.~the product group \emph{coats} consists of several product instances, which can be parkas, jackets, or other coats). Those semantic groupings constitute a hierarchical system (all \emph{coats} also belong to the more general group \emph{clothing}).

In an operational recommender system, we train the recommender model on six months (180 days) of user sessions. Our standard training set of six months of user sessions contains about 98.4 million sessions with 1.3 million distinct products, of which we consider the 1.1 million products with a frequency of at least 5. There are 850 product groups, and each group contains 1,587.85 products on average. We update the system on a daily basis, which entails that the source data shifts by one day. While guaranteeing up-to-date recommendations and computational tractability, this method produces a permanent instability factor. Although we retain 99\% of the data on each re-training, the results can differ considerably.

\section{How to measure instability}\label{experiment}
We consider instability of an algorithm as the sum of unexpected changes, assessable in prediction changes of the resulting model.
To measure instability, we thus need to be able to compare two models by quantifying and qualifying their differences. In the following, we describe the \emph{metrics} to accomplish this kind of analyses, which are applied in the \emph{experimental setup} outlined afterwards. 

\subsection{Metrics}\label{metrics}

We compare two click embeddings (trained on either two identical or modified data samples) using two types of metrics. 
One type  is application-driven and complies directly with changes in recommendations. For this measurement, we measure the \emph{neighborhood similarity} of the product vectors.
The other type of metric assesses the geometric differences in the vector model itself and analyzes \emph{the topological stability} of the manifolds constituted of the product vectors. 

\textbf{Neighborhood similarity:} To compute neighborhood similarity of two vector models, we sample some vectors and compare their nearest neighbors according to cosine similarity in both models. 
The top~{\it k} most similar vectors for a seed product represent the top~{\it k} best recommendations for the seed product in our use case.
For a given seed product~{\it p} we define the top~{\it k} overlap of two models as the fraction of the intersecting top~{\it k} recommendations predicted by both models.
In our experiment, we average over several pairwise model overlap scores, considering the intersection of a reference model and one test model at a time. For example, a top~{\it 15} overlap of 80\% for a given seed product and a reference model means that on average, 12 of 15 recommendations predicted by the reference model are also predicted by the test models.
Inversely, the same result quantifies instability as a difference of 20\%. 
In our final evaluation, we compute the overlap for the 15 nearest neighbors, averaged over 5,000 seed products randomly sampled among the 10,000 most frequent products. While this excludes rare products, this sample reflects the instabilities we will mostly see in practice.


\textbf{Topological stability:}  We complement the application-oriented neighborhood similarity with a model-driven measure which directly describes the geometric configuration of the vector model. As a tractable way to investigate properties and changes in vector constellation, we use a topological cluster analysis. Clusters arise from vectors according to the cosine similarity computed with the vector model.
To determine the number of connected components, we use the DBScan clustering algorithm \cite{Ester:1996:DAD:3001460.3001507}. The parameters for DBScan (neighborhood distance and the minimum number of neighbors) were heuristically determined and fixed for all runs. We require a core point to have at least ten neighbors with a cosine similarity $> ~0.8$. For our purpose, we do not require an optimal clustering, we just need a setting that reflects model changes deterministically.

To measure topological stability, we compute the number of connected components that is equivalent to the number of clusters. Further, we analyze the size and purity of clusters as well as the total amount of noise.

Additionally, we measure \emph{local density} in the vector space, counting the number of vectors within a certain distance from the clusters' centroids. We set this radius threshold to the value of $\varepsilon$-Parameter used for DBScan (0.8). To keep this analysis computationally tractable, we restrict the analysis to the 200 closest neighbors.

\subsection{Experimental Setup}\label{exp_setup}
We want to identify and separate random factors from minimal changes in input data. In order to keep the algorithm computationally tractable, we compile a comparably small and representative data sample consisting of 1,309,907 user sessions (2 consecutive days). The sample contains 503,936 distinct product types. Ignoring products that occur less than 5 times ({\it --min-count~5}), the resulting word2vec output contains 253,889 product types. On average, one session consists of 6.93 product clicks, and the product tokens in one session belong to 2.26 distinct product groups.

The model computed from the whole sample dataset is our \emph{reference model}.
We analyze the impact of a single data point (cf. Sec.~\ref{instability:data}) with a leaving-one-out experiment.
Based on the data sample described above, we create 500~subsamples by randomly leaving out one single session. For each subsample, we compute a vector model which we refer to as $model_{-1}$. The 500 sessions under consideration are representative with respect to session length (7.29 clicks). 

We compute the product embeddings using Google's word2vec \emph{WORD VECTOR estimation toolkit}, version 0.1c. 
For our sensitivity analysis towards minimal data changes, we need to eliminate algorithm-induced instabilities, which requires some parameter settings deviating from commonly used defaults. We choose skip-gram ({\it --cbow~0)}, no subsampling of frequent words ({\it --sample~0}), ten~iterations ({\it --iter~10}) and only one~thread due to thread-induced instabilities explained in Section~\ref{rand_factors}. We stick to 100 dimensions ({\it --size~100}) and calculate the word vectors for all 501~samples in two runs: One using Hierarchical Softmax (later on referred to as HS with {\it --hs~1 --negative~0}) and one with negative sampling with 5~samples (later on referred to as NEG with {\it --hs~0 --negative~5}).
Additionally, we fix the context window size to exactly~5 and round all vector numbers to 4 rather than 8 digits (the latter shows no impact in our results).

\section{Factors of Instability}\label{instability}
We analyze two types of instability. We consider the \emph{instability due to random factors} and describe the \emph{influence of the approximation method}. To disentangle artifacts from model training from actual sensitivity to input changes, we design an experiment that measures the prediction models' \emph{instability due to minimal data changes}.

\subsection{Random Factors out of the box}\label{rand_factors}
Using word2vec, we have learned that some seemingly random instabilities occur even for very simple, often small, datasets where, for instance, a single token that belongs to two sessions might introduce an opposing force that destabilizes the learned embedding. 
While our specific results are computed using Google's word2vec \emph{WORD VECTOR estimation toolkit v 0.1c}, many findings probably carry over to other embedding algorithms, too. 

Initially, we suspect random weight initialization in word2vec's code to account for most of the resulting variation. However, the word2vec implementation employs a number generator for each thread with a fixed seed and thereby guarantees reproducibility.

Analyzing other potential sources of randomness, we find one hyperparameter that induces a considerable amount of instability in the resulting word vectors: The \emph{number of threads}.

Using multiple threads for training is vital for large datasets, allowing to distribute all data equally over all threads. Since all threads update the same weight matrix without locking, the resulting weight updates are non-deterministic. To verify the impact, we train five models on the same input file with 30~threads\footnote{Additional parameters: -window 5 -size 100 -sample 0 -min-count 5 -cbow 0 -hs 1 -negative 0 -iter 10}. The models' agreement using similarity of the five nearest neighbors (cf. Sec.~\ref{metrics}) results in an average overlap of 75\% .

Another factor is \emph{subsampling} that randomly omits single tokens from sessions with respect to the frequency of tokens, controlled by the hyperparameter {\it --sample}. Furthermore, the definition of the context window size (that is at max {\it --window}) depends on the same random numbers as the subsampling.

While the subsampling can be deactivated in v 0.1c, there is no hyperparameter to keep the window size constant. When keeping the input data identical, these two factors will have no ``random'' effect on the output vectors.

\subsection{Approximation Methods for Model Training}
We use an algorithm that computes word embeddings from sequential data for all types in a vocabulary.
In their most basic count-based form, such embeddings treat each type as a dimension. 
With a growing vocabulary,  training time increases considerably, because there are more vectors to be updated with each processed training token.

In order to keep embedding computation tractable and maintain the resulting predictions' quality, many algorithms use approximation methods, which vastly influence the learned vector model as well as its sensitivity towards changes in the training data.
The word2vec implementation we use comes with the built-in options to either apply negative sampling (NEG), or Hierarchical Softmax (HS).

\textbf{Negative Sampling:} NEG saves runtime by reducing the number of vectors updated for each training token. The number is a new hyperparameter $z$. Instead of touching each vector in the vector space, only $z$ vectors (and additionally the vector for the current training token) will change at a time. The vectors are chosen using some probability distribution~$D$. $D$  can, for instance, be set to be the unigram distribution, which ensures that vectors of frequent concepts are updated more often.
\citeauthor{DBLP:journals/corr/MikolovSCCD13} show that this approach does not directly maximize the likelihood of correct tokens, but nevertheless it has been shown that this leads to useful embeddings and is thus a popular choice.
In the implementation of word2vec that we use, $z$ is by default set to 5. 

\textbf{Hierarchical Softmax:} HS is inspired from \citeauthor{Morin+al-2005} (\citeyear{Morin+al-2005}) and builds a binary tree
to reduce the complexity from $O(n)$ to $O(log(n))$, with $n$ being the number of types in the training corpus. Assuming e.g. one million of types, building the model requires only $log(n) \sim 20$ computation steps instead of $n=10^6$ computations.
The tree itself can be build in various ways. One common option is a Huffman tree that leads to binary codes where the
length of each code is proportional to the frequency of the given type. In this tree, each node represents a unique
type from the vocabulary with a unique code. 
Vector updates follow an order derived from root-to-leaf paths in the Huffman tree.

\textbf{Choice of Approximation Method:} Recent literature reflects NEG as the more popular option, but since more stable embeddings for recommendations are favored in practice, we nevertheless decide to use HS.
We study the difference of both sampling methods by evaluating how the resulting embeddings react to minimal data change with the experiment outlined in Sec.~\ref{experiment}. 
The correlation between the results of HS and NEG is about 50\%.
We find that NEG shows lower variance, but produces  more sensitive models with respect to both environment and data changes.

\begin{table}[!h] \centering
\centering
\small
\begin{tabular}{cp{0.7cm}p{0.79cm}p{1.0cm}p{1.0cm}c}
\toprule[1.5pt]
Pr. & F1 &  F2 & Code 1 & Code 2 &  Dist. \\ \midrule
A       & 15               & 15               & 01          & 10          & 2                \\ \midrule
B       & 10               & 10               & 00          & 00          & 0                \\ \midrule
C       & 9                & 9                & 110         & 110         & 0                \\ \midrule 
D       & 8                & 8                & 101         & 011         & 2                \\ \midrule
E       & 6                & 6                & 1111        & 1111        & 0                \\ \midrule
F       & 4                & 4                & 1001        & 1110        & 3                \\ \midrule
G       & 4                & 3 (-1)           & 1110        & 0100        & 2                \\ \midrule
H       & 3                & 3                & 10001       & 01011       & 3                \\ \midrule
I       & 1                & 1                & 10000       & 01010       & 3                \\ \bottomrule[1.5pt]
\end{tabular}
\caption{An example assuming two datasets (1 \& 2) with 9 products (Pr), their frequencies (F), Huffman codes \& Hamming distances.}
\label{dataset_huffman_codes}
\end{table}

\textbf{Hierarchical Softmax and Instabilities:} For our experiments that require a deterministic approximation method, we modify HS's tree building mechanism by processing types in their lexical order whenever they have the same frequency. 

While this modification guarantees identical prediction models for identical input, the tree building algorithm still accounts for much of the resulting models' sensitivity to seemingly inconsequential data changes: 
 Given that our input data follows a Zipf distribution, many of our types have similar frequencies, especially the long tail of rare products. If one single token of a type $t$  is omitted from the input (which means one click less), many leaves shuffle around and large parts of the inner tree structure can change. Each leaf node whose type shares $t$'s original frequency and each type node with $t$'s new frequency might change their position.

We illustrate such changes in the tree with a minimal example:
Table \ref{dataset_huffman_codes} shows the Huffman codes (Code 1 and 2) arising from an artificial setting of 9 product types (Pr. A--I). We compare two settings (1 and 2) that differ only by one occurrence of $G$ (in practice, this means a difference of one click). The Hamming distances ($Dist.$) between the two Huffman codes are shown in the last column.
In this example, decrementing the frequency of one type by one results in an average Hamming distance of 1.67.

When we scale this experiment to our sample data with 1,309,907 million sessions and 253,889 different products, one might assume that leaving out one data point becomes a trivial change. Leaving out one session from  1,309,907 does, however, impact the Huffman coding considerably: Of the 500 subsamples that differ by one session, only 13 share the same set of Huffman codes (and thus the same tree) with the reference model. Within the 487 models that have a different Huffman coding, on average the codes of 46,244$\pm$59,015 products change in a range between 3 and 244,259 changed codes. As we will show in the next section, changes in the Huffman tree directly affect the resulting vector model, and thus the final predictions.

\subsection{Results: Minimal Data Changes}\label{instability:data}
We evaluate the impact of single data points (user sessions) in two ways. First, we analyze the influence on neighborhood similarities in detail (cf. Section \ref{metrics}) using a linear regression model. This experiment quantifies associations between particular data point characteristics and the degree of overlap to the reference model. Additionally, we analyze topological changes via cluster analysis, which indicates how the model shape reflects sensitivity to missing data points.

\subsection*{Data Point Features \& Neighborhood Similarity}
We compare the overlap of the 15 nearest neighbors for 5,000 vectors sampled from the 10,000 most frequent types. The average overlap is 93.6$\pm$2.65 for HS and 80.5$\pm$1.98 for NEG.
With this measure as dependent variable, we analyze \emph{stability} rather than instability, and hence draw the reverse conclusion. We consider a model sensitive or unstable, if omitting a data point results in a new $model_{-1}$ that has only a small overlap with the reference model. As sampling with frequent types might be suspected to induce a biased view on the overlap, we validate this assumption drawing a random sample of types. This decreases the level of overlap slightly, but due to an almost perfect correlation to the used variable, analytical relations are virtually unaffected.

Table \ref{sessmin1_features} describes the explaining variables.
From that set, session \emph{Length} is the only property inherent to a data point, all other features describe data points in the training corpus' context (\emph{Frequency}, \emph{Rank}, \emph{Min Count}). Associated with the sampling method, we also count the number of changes in the \emph{Huffman codes} induced by leaving out this session (log10-scaled for a better fit), and the maximum change in binary code as quantified by the \emph{Hamming distance}.

\begin{table}
\centering
\small

\begin{tabular}{p{1.0in} p{1.9in}}
\toprule
Feature 							& Description 		    				\\ \midrule
Length        &  Number of products in session\\ 
\cellcolor{white}\cellcolor{lgray}Frequency & \cellcolor{lgray} Average frequency in the vocabulary scaled by 1,000\\ 
Rank  					&  Position in relative order in which the session is processed during training (scaled by $10^6$)\\
\cellcolor{lgray} Min Count &\cellcolor{lgray} Boolean: Omitting the session means at least one type falls under min count threshold\\
Huffman Codes & Number of changed codes due to leaving out session (Log10)\\
\cellcolor{lgray} Hamming Distance & \cellcolor{lgray}Maximum change in binary code as quantified by Hamming distance\\ 
\bottomrule

\end{tabular}
\caption{Features for single data points (user sessions).}\label{sessmin1_features}
\end{table}

Our analysis also compares both approximation methods, Hierarchical Softmax (HS) and Negative Sampling (NEG), verifying results across both methods. Except for \emph{Rank}, the features were selected such that the explanatory power for the HS-method is optimized; for the NEG-method choosing a different set of features would raise the degree of explicability.
Table \ref{sessmin1_models} summarizes the outcome:
\emph{Constant} shows the intercept of the regression model, \emph{Observations} is the number of considered $model_{-1}$-experiments and \emph{Adjusted R$^2$} displays the adjusted coefficient of determination.
                
According to the regression model’s results, the features account for about 92$\%$ of the variance in overlap (cf. Adjusted $R^2$) for HS, and about 67\% for NEG. For the HS-model, the two measures reflecting changes in the Huffman tree are by far the most important ones. They alone explain 85\% of the variation in the overlap, whereby stability is associated with fewer changes in the tree. 
A replicated leaving-one-out experiment with exactly the same Huffman tree for all runs shows this assumption to be true resulting in an average overlap of more than 99$\%$.

Interestingly, both Huffman-related measures also show a weakly significant correlation with NEG-overlap, although Negative Sampling does not employ Huffman trees. Exploring this  will be subject to future work.

Other relevant features include the session \emph{Length}, which correlates negatively with the overlap. Its significance is, however, bounded compared to changes in the Huffman tree. In a univariate regression model, session length alone can explain about 35\% of the variation. This relates to findings on linguistic data \cite{TACL1202}, with both results confirming the expectation that larger data chunks (documents or sessions) contain more information.

\emph{Frequency} is positively associated with overlap for HS. Omitting sessions with rare products decreases stability. Interestingly, rare items seem to affect NEG inversely.

Another feature, which has opposite associations for the two models, is \emph{Min Count}. This binary variable indicates whether one or more products fall under the threshold for consideration. While it seems odd that the stability of HS increases if products are left out, the association is weakly significant.

One feature exclusively relevant for NEG is the position of the left-out session in the input data file (\emph{Rank}). Sessions processed later have less influence on the models' stability.

\begin{table} \centering 
\small
\begin{tabular}{lrr} 

 \toprule
 Feature & HS & NEG\\ \midrule
 Length & $-$0.077$^{***}$ (0.004)  & $-$0.072$^{***}$ (0.005) \\ 
  \rowcolor{lgray}Frequency & 0.248$^{***} (0.059)$ & $-$0.487$^{***}$ (0.091) \\ 
  Huffman Codes & $-$1.200$^{***}$ (0.077) & $-$0.305$^{**}$ \hfill\hfill (0.120) \\ 
  \rowcolor{lgray}Hamming Distance & $-$0.160$^{***}$ (0.019) & $-$0.057$^{*}$\hfill\strut\hfill\hfill (0.030) \\ 
  Min Count & 0.239$^{*}$ \hfill(0.142) & $-$1.470$^{***}$ (0.222) \\ 
  \rowcolor{lgray}Rank & 0.004\hfill\strut\hfill (0.009) & 0.280$^{***}$ (0.014) \\ 
  Constant & 96.100$^{***}$ (0.279) & 80.300$^{***}$ (0.436)\\ 
 \hline \\[-1.8ex]
\rowcolor{lgray} Observations & \multicolumn{1}{c}{500} &  \multicolumn{1}{c}{500}  \\
 Adjusted R$^{2}$ & \multicolumn{1}{c}{0.92} & \multicolumn{1}{c}{0.67} \\
\bottomrule &&\\
 \multicolumn{3}{l}{\textit{Level of significance:} \textsuperscript{*} p$<$0.1; \textsuperscript{**}p$<$0.05; \textsuperscript{***}p$<$0.01} \\
\end{tabular}   \caption{Correlation coefficients of the linear regression model of data point features with the dependent variable: Overlap of 15 nearest neighbors. Standard Error in brackets.} 
\label{sessmin1_models} 
\end{table}

\subsection*{Topological stability}

We complement the analysis of neighborhood similarity with a study of model topologies.
While the overlap of recommendations quantifies a change with respect to a reference model, topological figures might allow to directly
benchmark models using their formal representations.

\begin{figure}
    \begin{centering}
            \includegraphics[width=200 pt]{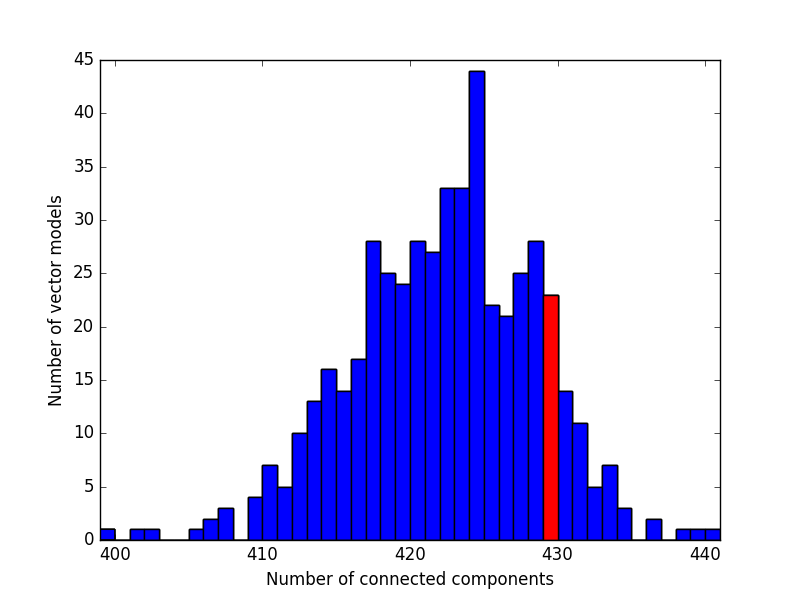}
            \caption{Number of clusters of the 501 click embeddings trained with HS. The bar with the reference model is marked in red.}
            \label{fig:connected_components}
    \end{centering}
\end{figure}

As an example, we illustrate the impact of a single data point on the number of clusters, which we calculate with the DBScan clustering algorithm (cf. Section \ref{metrics}).
Figure~\ref{fig:connected_components} shows a histogram of the 501 embeddings trained with the Hierarchical Softmax approximation. Each bar represents the number of $model_{-1}$-experiments (y-axis) resulting in the specific number of clusters (x-axis). 
The number of clusters ranges from 399 to 441 and the bar marked in red includes the reference model consisting of 429 clusters. The mean difference in the number of clusters with respect to the reference model is -7.23$\pm$6.21. In most cases, omitting a session reduces the number of clusters. 
Interestingly, the total number of noise points, which averages to 143,416$\pm$99, is virtually constant. Associated therewith, the average cluster size increases with a decreasing number of clusters, because the same number of points is divided into fewer clusters. 

We also analyze the purity of clusters, which we define as the proportion of products belonging to the clusters' prevalent product group. Purity is thus minimal if each product in a cluster belongs to a different group, and maximal if all products in a cluster belong to the same group.
The mean cluster purity for each embedding ranges from 0.903 to 0.917 with a mean of 0.910$\pm$0.002. The average purity depends on the number of small clusters, which contain fewer product groups and are thus purer in our experiment.

Tracing back the relationship of topological structure and model stability, we find mainly two bridges.  First, differences in the number of clusters are significantly larger for two models with large maximum Hamming distances and many Huffman code changes, which is in line with the findings of our leaving-one-out experiment.
Second, local density of the trained embeddings (cf. Section \ref{metrics}) correlates with the overlap measure. The larger the density of the embedding, the smaller the overlap. An intuitive explanation here is that there are many good recommendation candidates in a dense cluster, so the choice can vary more without sacrificing quality.
Future research needs to show how exactly such topological changes relate to the omitted data points.

\section{Conclusion}\label{conclusion}
We showed an approach to sensitivity analysis of geometric predictive models, compiling metrics that measure both the topological stability of the vector model itself and the similarity of the output when changing parameters or leaving out single data points.
As a basis for such an analysis, we identified and eliminated sources of randomness inherent to word2vec.

\subsection{Main Contributions}
Our main take-aways might help others to interpret and prevent instabilities in word2vec and similar algorithms:
\begin{itemize}
\item \textbf{Stable settings:} We identified sources of instability in word2vec and showed how to achieve constant results on identical input data. In particular, we showed that fixing the Huffman tree for HS can prevent variance even when the input data changes.
\item \textbf{Choice of Approximation Method:} Choosing a different approximation method
changes both, the result of a model and its sensitivity. In word2vec's case, Negative Sampling adopts new information more readily, while Hierarchical Softmax maintains more stability.
\item \textbf{Data changes:} Even a single data point can change the models' output considerably. While the results' quality might be stable, the actual outputs change depending on the actual data points added or removed.
\item \textbf{Information Density:} Data points, which hold more information, cause more changes in the final model. ``More information'' does not only refer to longer sentences or user sessions, but in particular to data points containing surprising associations of concepts, or  information on rare concepts, or information on concepts for which the model learned a ``cluttered'' environment.
\end{itemize}

\subsection{Discussion \& Future Work}

We presented a detailed way to measure model and output stability on an industrial scale. While certainly not comprehensive, our metrics can be a starting point for further analysis on model sensitivity of embeddings. We also believe that our methods and findings are applicable to linguistic and other applications. However, language as a data source behaves in a considerably different fashion than user interactions, thus possible differences concerning expressiveness and usefulness of our way to measure output consistency and quality can be expected. Comparing our results applied with linguistic experiments could show interesting differences.

In future work, we aim to additionally relate topological metrics to data point features. In the long run, we want to find a generic way to detect and predict sensitivity for all types of vector models, independently from particular use cases, data types or scale. Another goal is to highlight the relationship of model instability and business value of certain data points, assuming that data points which increase the model's output quality carry more valuable information. With such experiments, we could show not only data-point influence on model structure and predictions, but also measure the monetary value of model stability and the underlying data.

\section*{Acknowledgments} We want to thank Julia Soraya Georgi and two anonymous reviewers for their helpful comments. All remaining errors are of course our own. 

\bibliographystyle{named}
\bibliography{ijcai18}

\end{document}